\documentclass[twocolumn,10pt]{article}

\usepackage[a4paper,margin=2cm,footskip=6mm]{geometry}
\usepackage{times}
\usepackage{booktabs} % For formal tables
\usepackage{cite}
\usepackage{amsmath,amsthm,amssymb,amsfonts,mathtools}
\usepackage{graphicx}
\usepackage{graphics}
\usepackage{textcomp}
\usepackage{xcolor}
\usepackage{layouts}
\usepackage{multirow}
\usepackage{tabularx}
\usepackage{multirow}
\usepackage{enumitem}
\usepackage{bm}
\usepackage[utf8]{inputenc}

\usepackage{algorithm}
\usepackage{booktabs}
\usepackage{etoolbox}
\usepackage{tabularx}
\usepackage{xparse}
\usepackage{algpseudocode}
\usepackage{caption}
\usepackage{subcaption}

\newcommand{\acc}{acc}
\newcommand{\batchSize}{b}
\newcommand{\nClasses}{C}
\newcommand{\uncFunc}{u}
\newcommand{\distLab}{dl}
\newcommand{\distUnlab}{du}
\newcommand{\learner}{\theta}
\newcommand{\maxLength}{M}
\newcommand{\nnSampleSize}{k}
\newcommand{\nnInputTuple}{T}
\newcommand{\nnInputBatchBatch}{B}
\newcommand{\randomActionSet}{\mathbb{U}}
\newcommand{\singleNeuronInput}{\mathcal{J}}
\newcommand{\preSamplingResult}{\mathcal{P}}
\newcommand{\preSamplingIterations}{j}
\newcommand{\labeledSet}{\mathcal{L}}
\newcommand{\predUnity}{pu}
\newcommand{\unlabeledSet}{\mathcal{U}}
\newcommand{\unlabeledQuery}{\unlabeledSet_q}

\newcommand{\MDPPolicy}{\pi}
\newcommand{\MDPStateSet}{\mathcal{S}}
\newcommand{\MDPActionSet}{\mathcal{A}}
\newcommand{\MDPCurrentState}{s}
\newcommand{\MDPCurrentAction}{a}
\newcommand{\MDPReward}{R}
\newcommand{\MDPProbability}{P}
\newcommand{\AmountOfSDs}{\alpha}
\newcommand{\AmountOfTrainingALCycles}{\beta}

\DeclareMathOperator*{\argmax}{argmax}
\newcommand{\ImitAL}{\textsc{ImitAL}}

\begin{document}

\date{}

\newcolumntype{C}[1]{>{\centering\arraybackslash}p{#1}}
\newcolumntype{R}[1]{>{\raggedleft\arraybackslash}p{#1}}
\newcolumntype{L}[1]{>{\raggedright\arraybackslash}p{#1}}

\MakeRobust{\Call}

\title{\Large \bf ImitAL: Learning Active Learning Strategies from Synthetic Data}

\author{
{\rm Julius Gonsior, Maik Thiele, Wolfgang Lehner}\\
 Technische Universität Dresden \\
Dresden, Germany\\
\textless firstname.lastname\textgreater @tu-dresden.de
}

\maketitle

\thispagestyle{empty}

\subsection*{Abstract}
	One of the biggest challenges that complicates applied supervised machine learning is the need for huge amounts of labeled data.
	\emph{Active Learning} (AL) is a well-known standard method for efficiently obtaining labeled data by first labeling the samples that contain the most information based on a query strategy.
	Although many methods for query strategies have been proposed in the past, no clear superior method that works well in general for all domains has been found yet.
	Additionally, many strategies are computationally expensive which further hinders the widespread use of AL for large-scale annotation projects.

	We, therefore, propose \ImitAL{}, a novel query strategy, which encodes AL as a learning-to-rank problem.
	For training the underlying neural network we chose \emph{Imitation Learning}.
	The required demonstrative expert experience for training is generated from purely synthetic data.

	To show the general and superior applicability of \ImitAL{}, we perform an extensive evaluation comparing our strategy on 15 different datasets, from a wide range of domains, with 10 different state-of-the-art query strategies.
	We also show that our approach is more runtime performant than most other strategies, especially on very large datasets.

\section{Introduction}
The quality of Supervised Learning depends inherently on the amount and quality of the labeled dataset.
Acquiring labels is a time-consuming and costly task that can often only be done by domain experts.
% \emph{Active Learning} (AL) is a standard approach for saving human effort by iteratively selecting only those unlabeled samples for labeling by their usefulness for the classification task.
\emph{Active Learning} (AL) is a standard approach for saving human effort by iteratively selecting only those unlabeled samples for labeling that are the most useful ones for the classification task.
The goal is to train a classification model $\learner$ which maps samples $x \in \mathcal{X}$ to a respective label $y \in \mathcal{Y}$.

\begin{figure}[t]
	\centering
	\includegraphics{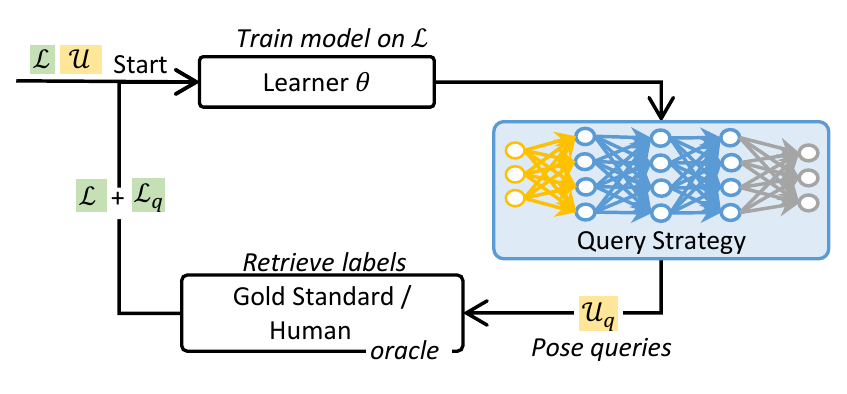}
	\caption{Standard Active Learning Cycle}
	\label{fig:al_cycle}
\end{figure}

% We consider pool-based AL sampling scenarios.
Figure~\ref{fig:al_cycle} shows a standard AL cycle for a pool-based sampling scenario, which we are focusing on.
Given a small initial labeled dataset $\labeledSet = \{(x_i,y_i)\}_i^n$ of $n$ samples $x_i \in \mathcal{X}$ and the respective label $y_i \in \mathcal{Y}$ and a large unlabeled pool $\unlabeledSet = \{x_i\}, x_i \not\in \labeledSet$ a learner $\learner$ is trained on the labeled set.
Afterward, the \emph{query strategy} chooses a batch of $\batchSize$ unlabeled samples $\unlabeledQuery$.
This batch gets labeled by the oracle and is added to the labeled set $\labeledSet$, and the whole AL cycle repeats until some stopping criteria are met.

During the past years, many different query strategies have been proposed, but to our knowledge, none excels consistently over a large number of datasets from different application domains.
Typical early query strategies include simple uncertainty-based variants~\cite{lc_sampling,mm_sampling,ent_sampling} or query-by-committee strategies~\cite{qbc_sampling}, which are still widely popular, despite a vast number of recently proposed more elaborate strategies~\cite{ALIL,LAL-RL,Pang_single,SPAL}.
Even though various empirical comparisons exist~\cite{settles_al_survey,deep_al_surley}, no clearly superior strategy could be found.
The results are mixed and suggest that current AL strategies highly depend on the underlying dataset domain.
Often even the na\"ive baseline of randomly selecting samples achieves surprisingly competitive results~\cite{LAL-RL,Pang_single,ALIL,QUIRE,ALBL,BMDR}.
In addition to that, practical implications like high computational costs of several strategies prevent them from being used in large-scale annotation projects.

We propose therefore \emph{\ImitAL{}}\footnote{URL to code repository omitted due to anonymization}, a novel approach of learning an AL query strategy using \emph{Imitation Learning} from purely synthetic datasets that performs well independently of the classification domain with efficient runtime performance, even for large-scale datasets.
% By efficient runtime performance we mean the amount of time needed for the query strategy to calculate its output.
The task of selecting the most informative unlabeled samples is treated as a listwise learning-to-rank~\cite{Listwise} problem and is solved by an artificial \emph{neural network} (NN).
The training of the NN is performed in a novel way of generating a large number of synthetic datasets and calculating greedily an optimal label ranking.
Thereafter, the ranking is used as an expert strategy to train the NN in an imitation learning setting.
The most difficult task thereby is the encoding of the unlabeled samples for the NN, while still complying with the requirement of being independent of the characteristics of the dataset.
Even though true generalization is nearly impossible to prove, we base our assumption of universal applicability of \ImitAL{} likewise to~\cite{LAL-RL,Pang_single}.
Our proposed method does not rely on the underlying AL learner model, the classification domain, or the feature space of the dataset and can therefore be used in any supervised machine learning project during the labeling phase.
In contrast to many other learned AL methods~\cite{PAL,WoodwardandFinn,ALIL,LAL-RL,Bachman,Pang_single}, we provide a readily applicable AL query strategy NN which does not require an initial step of transfer learning on domain-related datasets before being applicable.
Under the assumption that our NN is trained on a large set of synthetic datasets, which resembles the set of possible real-world datasets, it seems to be universally applicable to any possible dataset, independent of the domain.
Another big difference to other AL methods is the small runtime of \ImitAL{}, especially on very large datasets.
% Thus, our trained AL query strategy can be already used in all domains and does not need an initial transfer-learning phase on domain-related datasets.
%
%
\begin{figure}[t]
	\centering
	\includegraphics[width=\columnwidth]{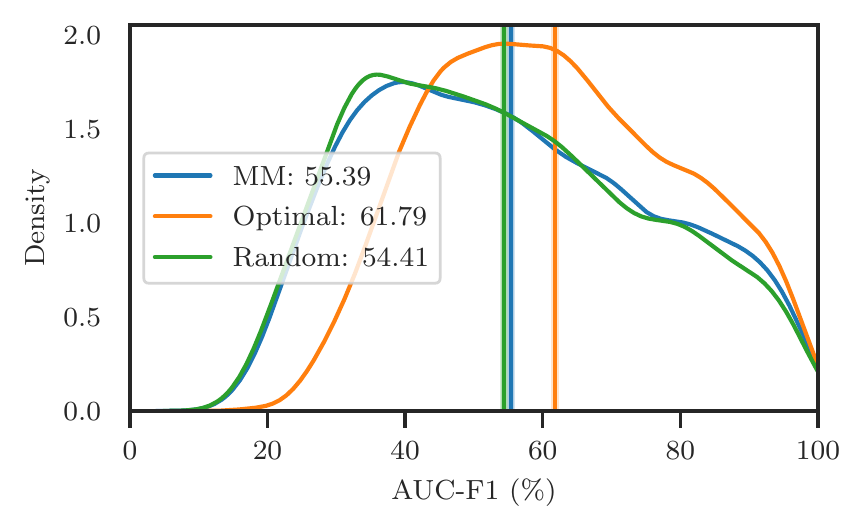}
	\caption{Possible improvements by a greedy omniscient AL strategy to uncertainty max-margin (MM) and random selection}
	\label{fig:optimal}
\end{figure}
\paragraph{Outline.} The remainder of this paper is organized as follows:
In Section~\ref{sec:MDP} we formally define the encoding of the AL query strategy problem as a Markov Decision Problem as well as the imitation learning phase, and how the encoding of the unlabeled samples as input for the NN works.
Section~\ref{sec:train_imital} illustrates the practical training phase from fully synthetic datasets, and Section~\ref{sec:evaluation} an extensive evaluation of different state-of-the-art query strategies over multiple real-world datasets from different domains.
Finally, we present related work in Section~\ref{sec:related_work} and conclude in Section~\ref{sec:conclusion}.
\section{Active Learning as Markov Decision Problem}
\label{sec:MDP}
At its core, the AL query selection can be seen as a \emph{Markov Decision Problem} (MDP).
An MDP is a 4-tuple $(\MDPStateSet,\MDPActionSet,\MDPProbability_\MDPCurrentAction, \MDPReward_\MDPCurrentAction)$ containing the set of states $\MDPStateSet$, the set of actions $\MDPActionSet$, the probability $\MDPProbability_\MDPCurrentAction(\MDPCurrentState, \MDPCurrentState') = Pr(\MDPCurrentState_{t+1}=\MDPCurrentState' | \MDPCurrentState_t = \MDPCurrentState, \MDPCurrentAction_t = \MDPCurrentAction)$ that action $\MDPCurrentAction$ in state $\MDPCurrentState$ at time $t$ will lead to state $\MDPCurrentState'$, and the reward $\MDPReward_\MDPCurrentAction(\MDPCurrentState,\MDPCurrentState')$ received after applying action $\MDPCurrentAction$ to transition from state $\MDPCurrentState$ to state $\MDPCurrentState'$.
When defining AL as an MDP problem the state space consists of the labeled set $\labeledSet$, the unlabeled set $\unlabeledSet$, and the currently trained learner model $\learner$.
The action space consists of all possible batches of unlabeled samples $\unlabeledQuery \subseteq \unlabeledSet$ of length $\batchSize$.
The optimization goal of an MDP is to find a \emph{policy} $\MDPPolicy: \MDPStateSet \mapsto \MDPActionSet$ that selects the best action for a given state. %: $\MDPPolicy(\MDPCurrentState) = \argmax_{\MDPCurrentAction \in \MDPActionSet} \MDPReward_\MDPCurrentAction()$
For the case of AL that would be the query strategy.
Instead of manually defining a policy we train an NN to function as the policy for \ImitAL{}.
The input of the NN will be both the state and multiple actions, or in other words, multiple samples to label, and the output denotes which of the possible actions to take.
The AL query strategy is therefore implemented as an NN that solves a listwise learning-to-rank~\cite{Listwise} problem, ranking the set of unlabeled samples $\unlabeledSet$ to select the most informative batch $\unlabeledQuery$.

In Section~\ref{sec:il} we will first define the learning process of the policy NN via imitation learning, in Section~\ref{sec:state_space} the policy NN input encoding, and in Section~\ref{sec:action_space} the policy NN output encoding.
\subsection{Imitation Learning}
\label{sec:il}
We use imitation learning to train the NN functioning as policy.
Instead of formulating a reward function, an expert demonstrates optimal actions, which the policy NN attempts to imitate.
For AL it is easy to demonstrate a near-optimal greedy policy, but hard to manually define an optimal policy, or to formulate the reward function.
We use \emph{behavioral cloning}~\cite{Michie94buildingsymbolic}, which reduces imitation learning to a regular supervised learning problem.
Given a large set of states, the corresponding possible actions, and the optimal action, one can directly train a classification model using the possible actions as input and the indicating optimal actions as the target output.
\begin{algorithm}[t]
	\caption{Training process of \ImitAL{}}
	% \scriptsize{\textbf{Input:} small clustered hm labeled start set $\mathcal{L}$, large unlabeled clustered dataset $\mathcal{U}$, query strategy $QS$, batch size $BS$, human user interaction budget $B$, minimum training accuracy before WS $M$, minimum certainty threshold $\alpha$, minimum cluster homogeneity $\beta$, minimum labeled cluster size $\gamma$ and a cluster query strategy $CQS$}\\
	% \scriptsize{\textbf{Output:} labels for $\mathcal{U}$}
	\begin{algorithmic}[1]
		\State Init $\mathcal{I} = \emptyset; \mathcal{O} = \emptyset$ \Comment{Input and output for policy NN}
		\State $i=\AmountOfSDs$ \Comment{Amount of synthetic datasets to generate}
		\While{$i>0$}
			\State $\labeledSet, \unlabeledSet \gets \Call{GenerateSyntheticDataset}$
			\State $j=\AmountOfTrainingALCycles$ \Comment{Amount of AL cycles to use this dataset}
			\While{$j>0$}
				\State $\Call{train}{\learner, \labeledSet}$
				\State Init $accs = \emptyset$
				\State Randomly draw $\preSamplingIterations$ possible action sets $\{\randomActionSet_1, \dots, \randomActionSet_\preSamplingIterations\}$
				\State $\preSamplingResult= \{\singleNeuronInput_1, \dots, \singleNeuronInput_\nnSampleSize\} = \Call{PreSampling}{\randomActionSet_1, \dots, \randomActionSet_\preSamplingIterations}$
				\For{$\singleNeuronInput_i \in \preSamplingResult$}
					\State Get labels $\labeledSet_i$ for $\singleNeuronInput_i$
					\State $\Call{Append}{accs, \Call{acc}{\learner, \labeledSet \cup \{(\singleNeuronInput_i, \labeledSet_i)\}}}$
				\EndFor
				\State $\Call{Append}{\mathcal{I}, \Call{InputEncoding}{\randomActionSet, \labeledSet, \unlabeledSet, \learner}}$
				\State $\Call{Append}{\mathcal{O}, accs}$
				\State Take highest action $\unlabeledQuery$ based on $accs$
				\State Add labels of $\unlabeledQuery$ to $\labeledSet$
				\State Remove $\unlabeledQuery$ from $\unlabeledSet$
				\State $j \gets j-1$
			\EndWhile
			\State $i \gets i-1$
		\EndWhile
		\State $\Call{Train}{\MDPPolicy, \mathcal{I}, \mathcal{O}}$
	\end{algorithmic}
	\label{alg:trainImitAL}
\end{algorithm}
%
% We train a NN as policy or AL query strategy.
%
%
\begin{figure*}[t]
	\centering
	\begin{subfigure}{\linewidth}
		\includegraphics{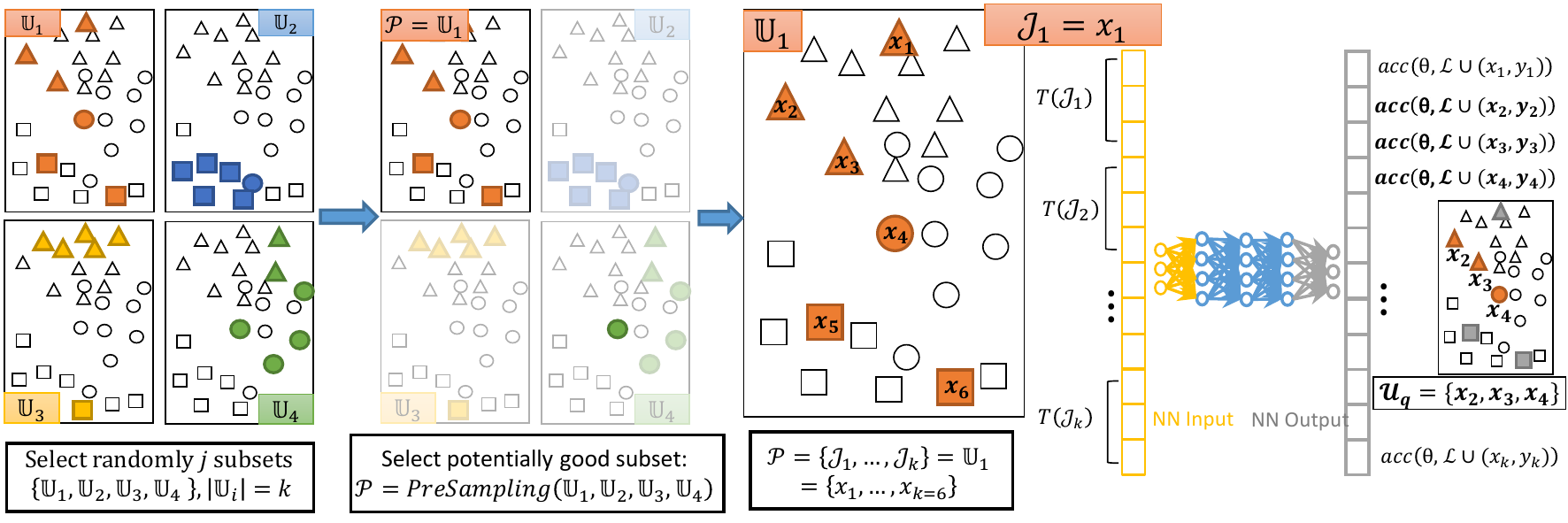}
		\caption{\ImitAL{} single variant, example for $\preSamplingIterations$= 4, $\nnSampleSize$=6, and  $\batchSize$=3}
	\end{subfigure}
	\begin{subfigure}{\linewidth}
		\includegraphics{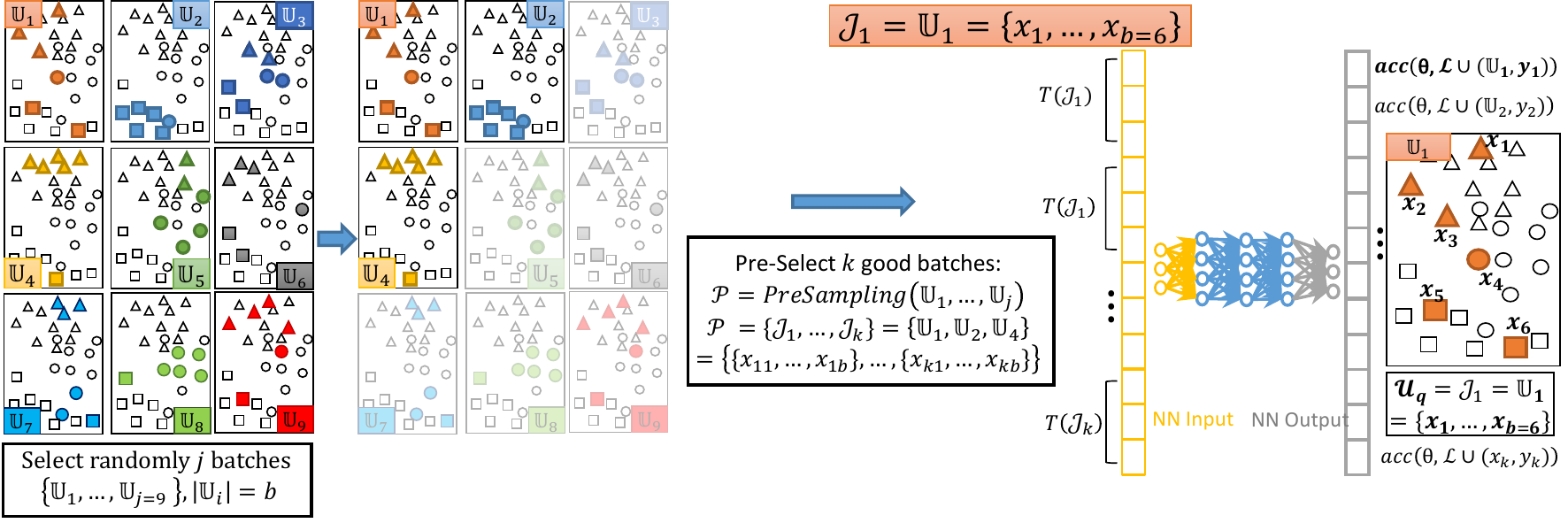}
		\caption{\ImitAL{} batch variant, example for $\preSamplingIterations$= 9, $\nnSampleSize$=3, and  $\batchSize$=6   }
		% \label{fig:nn_enc_batch}
	\end{subfigure}
	\caption{Input and output encoding for both variants of \ImitAL{}}
	\label{fig:nn_enc}
\end{figure*}

The general algorithm to train the policy NN of \ImitAL{} is detailed in Algorithm~\ref{alg:trainImitAL}.
Basically, the policy NN solves the ranking problem of selecting the batch $\unlabeledQuery$ of the $\batchSize$-best unlabeled samples out of the set of unlabeled samples $\unlabeledSet$.
The trained AL strategy by the policy NN should be applicable to all kinds of classification problems.
Therefore, in an ideal setting, one would first enumerate all possible datasets, and second compute the optimal order in which the unlabeled samples should be labeled, resulting in an exhaustive input-output list.
Since this is computationally not feasible, we make two assumptions to solve both problems: first, by approximating with a large number of synthetic datasets we can get close to the diverse range of real-world data sets and achieve near-universal applicability, and, second, we can select greedily the near-optimal samples per AL cycle since we know for the synthetic datasets the correct labels a priori.
% As both is obviously not possible, we approximate by a large number of synthetic datasets.
% Since we know for the synthetic dataset the correct labels a priori we can approximate the optimal samples per AL cycle.
For the second assumption, we perform a \emph{roll-out} into the next AL cycle, where we calculate the possible accuracy of the AL learner $\learner$ in the next AL cycle for applying the possible action $\singleNeuronInput \subseteq \unlabeledSet$ by $\acc(\theta, \labeledSet \cup \{(\singleNeuronInput_i, \labeledSet_i)\})$.
Since we are training on synthetic datasets, we already know the labels $\labeledSet_i$ for $\singleNeuronInput_i$ a-priori.
The policy NN AL query selection solves therefore the regression problem of predicting the expected resulting accuracy for each unlabeled sample, if one would label this sample.
Given the accuracies of the respective actions, we assume that those actions yielding higher accuracies are a far better policy than any standard AL strategies.
Figure~\ref{fig:optimal} shows the discrepancy between the greedy near-optimal AL strategy, random selection, and uncertainty max-margin, which, as we will show later in our evaluation in Section~\ref{sec:evaluation}, is one of the best performing state-of-the-art AL strategies.
Each displayed distribution contains the AUC-F1-Score after 50 AL cycles on 10,000 different synthetic datasets.
The metric is explained in detail in Section~\ref{sec:exp_setup_details}.
Note, that an AUC-F1-Score of 100\% is of course not always possible.
The gap between the uncertainty max-margin strategy and the greedy near-optimal actions indicates that there is lots of room for improvement.
Hence we argue that this greedy strategy is sufficiently good as an imitation learning expert.
One can also notice in this picture, that the range of possible improvement of AL strategies compared to the random strategy is rather small percentage-wise.
% As the regression target output worked better than the classification one, the AL query selection policy NN solves therefore the regression problem of predicting the expected resulting accuracy for each unlabeled sample, if one would label this sample.
As training data, we generate $\AmountOfSDs$ synthetic datasets and perform for each synthetic dataset $\AmountOfTrainingALCycles$ AL cycles to gather state-action pairs.
The resulting $\AmountOfSDs \cdot \AmountOfTrainingALCycles$ state-action pairs are then directly used to train the policy NN.

Note that in contrast to most traditional AL query strategies, we do not feed all possible actions out of the complete set of unlabeled samples $\unlabeledSet$ into the policy NN, but a set $\preSamplingResult$ of $\nnSampleSize$ potential actions $\singleNeuronInput$ only once each AL cycle $\preSamplingResult = \{\singleNeuronInput_1, \dots, \singleNeuronInput_\nnSampleSize\}, \singleNeuronInput \subset \unlabeledSet$.
This is due to the fixed input size of artificial neural networks and the varying number of actions for different datasets.
Another added benefit is that we only have to perform the future roll-out on the $\nnSampleSize$ possible actions, instead of all possible actions from $\unlabeledSet$ during the training phase.
We use a simple heuristic defined by the function $\Call{PreSampling{}}$ for selecting potentially good actions $\preSamplingResult$ first, instead of purely random selection, to compensate for this restriction.
Details of the heuristic are given in the following Section~\ref{sec:state_space}.
A positive side effect of the fixed-size input of the policy-NN is the low and static runtime of \ImitAL{}, which is almost independent of the sample size.
The effect is especially apparent with very large datasets.
\subsection{Policy NN Input Encoding}
\label{sec:state_space}
One of the most important parts of formulating AL as an MDP with an imitation learning policy is encoding the input and output state suitable for an NN policy.
Figure~\ref{fig:nn_enc} displays the general procedure of the encodings.
We propose two variants of \ImitAL{}, one, which selects individual samples, and the other one, which selects batches at once.
The main difference is the policy NN output encoding: for the single variant, each output neuron of the policy NN represents an action or a single unlabeled sample, for the batch variant each output neuron a complete batch of $\batchSize$ samples.
However, both variants are batch-aware in terms as of AL strategies as one can use a batch of the $\batchSize$ unlabeled samples indicated by the $\batchSize$-highest output neurons.
% Both variants select as actions complete batches of $\batchSize$ samples for the oracle, for the single variant are exactly $\batchSize$ output neurons taken as action, whereas for the batch variant one is sufficient.
Batch-awareness is, as thoroughly explained in~\cite{BatchBALD}, a beneficial and desirable property of AL query strategies, meaning that the joint diversity of all samples of the final AL batch $\unlabeledQuery$ is taken into account.
Also different is the calculation of the future roll-out when calculating the imitation learning expert, as for the single variant the future accuracy is calculated independently for each sample, and for the batch variant jointly per batch.
Therefore, even though the learned AL strategy is batch-aware, the imitation learning expert is only batch-aware for the batch variant of \ImitAL{}.

As noted before, we provide $\nnSampleSize$ input actions into the policy NN at once.
Therefore, we first need to select a potentially good set of $\nnSampleSize$ actions $\preSamplingResult=\{\singleNeuronInput_1, \dots, \singleNeuronInput_\nnSampleSize\}$.
For that, a set of $\preSamplingIterations$ unlabeled action sets $\{\randomActionSet_1, \dots, \randomActionSet_\preSamplingIterations\}$ is randomly drawn.
For the single variant $\singleNeuronInput$ represents a single unlabeled sample and each $\randomActionSet$ is a set of $\nnSampleSize$ unlabeled samples $\randomActionSet \subset \unlabeledSet, |\randomActionSet| = \nnSampleSize$.
The pre-sampling selection selects the most promising $\randomActionSet$ and uses it as $\preSamplingResult$.
Different for the batch variant, $\randomActionSet$ is directly a complete batch $\randomActionSet \subset \unlabeledSet, |\randomActionSet| = \batchSize$ and the pre-sampling selection selects the $\nnSampleSize$ most promising batches $\randomActionSet$ = $\singleNeuronInput$ directly as $\preSamplingResult$.
% For the single variant $\randomActionSet$ represents a set of single samples $\randomActionSet = \{\singleNeuronInput_1, \dots, \singleNeuronInput_\nnSampleSize\}, \singleNeuronInput \in \unlabeledSet$, and for the batch variant a set of complete batches $\randomActionSet = \singleNeuronInput, \singleNeuronInput  \subset \unlabeledSet, |\singleNeuronInput| = \batchSize$.
% The function $\Call{PreSampling}{}$ is used to select the best actions $\preSamplingResult$ out of the $\preSamplingIterations$ random ones.
The complete input of the policy NN is then an encoding of the actions, which also contains parts of the state, resulting in a set of tuples: $\Call{InputEncoding}{\preSamplingResult} = \{\nnInputTuple(\singleNeuronInput) | \singleNeuronInput \in \preSamplingResult\}$.
% We select randomly $\preSamplingIterations$-times possible actions and select out of those the $\nnSampleSize$-potentially best ones according to the heuristic.
% This also accelerates the training process as potentially better 
% To accelerate the training process we compensate this restriction by a simple heuristic defined by the function $\Call{PreSampling}{}$ for selecting potentially good samples first, instead of pure random selection.
After the policy NN is trained, this simple heuristic is not needed anymore for applying, as the NN itself functions better than the heuristic, and works well with a pure random subset as input.
The following two sections describe the respective definition of $\nnInputTuple(\singleNeuronInput)$ and $\Call{PreSampling}{}$.
% and for the batch variant: $\Call{InputEncoding}{\nnInputBatchBatch_1, \dots, \nnInputBatchBatch_\nnSampleSize} = \{\nnInputTuple_{batch}(\nnInputBatchBatch_1), \dots,  \nnInputTuple_{batch}(\nnInputBatchBatch_1)\}$
% \\
% , for the single variant $\nnSampleSize$ unlabeled samples $\hat{\unlabeledSet_\nnSampleSize} \subseteq \unlabeledSet$, and for the batch variant $\nnSampleSize$ $\batchSize$-length batches of samples: $\nnInputBatchBatch = {x_1, \dots, x_\batchSize}, x_i \in \unlabeledSet$.
% After that, using the input encoding, each sample $x_i$ gets encoded either by the single encoding function $\nnInputTuple_{single}(x)$ or a complete batch of $\batchSize$ samples $x_i$ by the batch encoding function $\nnInputTuple_{batch}(x_1, \dots , x_\batchSize)$.
% The complete input of the policy NN for the single variant is a set of $\nnSampleSize$ tuples: $\Call{InputEncoding}{\hat{\unlabeledSet_k}} = \{\nnInputTuple_{single}(x) | x \in \hat{\unlabeledSet_\nnSampleSize}\}$ and for the batch variant: $\Call{InputEncoding}{\nnInputBatchBatch_1, \dots, \nnInputBatchBatch_\nnSampleSize} = \{\nnInputTuple_{batch}(\nnInputBatchBatch_1), \dots,  \nnInputTuple_{batch}(\nnInputBatchBatch_1)\}$
% %
%
%
\subsubsection{Single Input Encoding}
\label{sec:singleInput}
% The input of the policy NN represents the set of unlabeled samples $\unlabeledSet$.
The single input variant works on encoding individual samples $\singleNeuronInput \in \unlabeledSet$.
% The pre-sampling function $\Call{PreSampling}{}$ returns a subset $\hat{\unlabeledSet_\nnSampleSize} \subseteq \unlabeledSet$ of length $\nnSampleSize$ of unlabeled samples $x$ as possible actions.
For the training of \ImitAL{} we are using a simple heuristic $\Call{PreSampling}{}$ to filter out potentially uninteresting actions, whereas for the application after the training, pure random selection works just fine.
We calculate the average distance to the already labeled samples of all the samples in each possible action set $\randomActionSet_i$ and select the action set $\preSamplingResult$ having the highest average distance:
% For a training, we randomly select $\preSamplingIterations$ times $\nnSampleSize$ unlabeled samples, calculate the average distance of the $\nnSampleSize$ samples to the already labeled samples, and select those $\nnSampleSize$ unlabeled samples out of $\preSamplingIterations$, having the highest average distance to the labeled samples:
\begin{align}
	\preSamplingResult = \Call{PreSampling}{\randomActionSet_1, \dots, \randomActionSet_\preSamplingIterations} = \argmax_{\randomActionSet_i} \sum_{x \in \randomActionSet_i}\distLab(x),
\end{align}
% where $\randomActionSet_i$ contains $\nnSampleSize$ unlabeled samples.
Thus, we are ensuring that we sample evenly distributed from each region in the sample vector space during the training process.

The input encoding $\nnInputTuple_{single}(x)$ defines on what basis the policy can make the AL query strategy decision.
A good AL query strategy takes informativeness as well as representativeness into account.
The first favors samples, which foremost improve the classification model, whereas the latter favors samples that represent the overall sample distribution in the vector space.
Informativeness is derived by $\uncFunc_i(x)$, a function computing the uncertainty of the learner $\learner$ for the $i$-th most probable class for the sample $x\in\unlabeledSet$, given the probability of the learner $P_{\learner}(y|x)$ in classifying $x$ with the label $y$:
\begin{align}
	\uncFunc_i(x) =
	\begin{cases}
		P_{\learner}\Bigl(\left(\argmax_{y,i} P_\learner(y|x)\right)\Bigm|x\Bigr), & \text{if } i \leq \nClasses \\
		0,                                                                         & \text{otherwise}
	\end{cases}
\end{align}%
Note that $\argmax_{\_~,i}$ denotes the $i$-th maximum argument, and $\nClasses$ the number of classification classes.
The well-known uncertainty least confidence AL query strategy would be $\uncFunc_1(x)$, and uncertainty max-margin $\uncFunc_1(x)-\uncFunc_2(x)$.

For representativeness we compute $\distLab(x)$ and $\distUnlab(x)$, the first denoting the average distance to all labeled samples, the latter the average distance to all unlabeled samples:
\begin{align}
	\distLab(x)   & = \frac{1}{|\labeledSet|} \sum_{x_l \in \labeledSet} d(x,x_l)      \\
	\distUnlab(x) & = \frac{1}{|\unlabeledSet|} \sum_{x_u \in \unlabeledSet} d(x,x_u),
\end{align}%
where $d(x_1,x_2)$ is an arbitrary distance metric between point $x_1$ and point $x_2$.
We use the Euclidean distance for small feature vector spaces, and recommend using the cosine distance for high-dimensional feature vector space.
This is discussed in detail in Section~\ref{sec:evaluation}.

The final input encoding for an action $\singleNeuronInput$ or sample $x$ is the five-tuple:
\begin{align}
	\nnInputTuple_{single}(\singleNeuronInput = x) = \left(\uncFunc_1(x), \uncFunc_2(x), \uncFunc_3(x), \distLab(x), \distUnlab(x)\right)
\end{align}
\subsubsection{Batch Input Encoding}
\label{sec:batchInput}
In contrast to the first variant of \ImitAL{}, the batch variant uses a complete batch of $\batchSize$ samples $x$ as actions $\singleNeuronInput \subset \unlabeledSet$ using the encoding:
\begin{align}
	\nnInputTuple_{batch}(\singleNeuronInput = \{x_1, \dots , x_\batchSize\}) = \begin{pmatrix}
		\frac{1}{\batchSize} \sum_{i=1}^\batchSize \uncFunc_1(x_i),                                 \\
		\frac{1}{2 \batchSize \maxLength } \sum_{i=1}^\batchSize \sum_{j=1}^\batchSize d(x_i, x_j), \\
		\predUnity(x_1, …, x_\batchSize) \vphantom{\sum_{i=1}^b}
	\end{pmatrix}
\end{align}%
The first part of the batch encoding $\nnInputTuple_{batch}$ is the average uncertainty of the samples in the batch.
The second part is the average distance between all samples in the batch, normalized with the maximum possible distance $\maxLength$ in the vector space of the samples.
The last part is \emph{predicted unity} $pu$ of a batch of samples, which takes the predicted classes of the underlying learner $\learner$ as input, and calculates a percentage agreement score between them, 1 means complete agreement, and 0 complete disagreement.
All three components of $\nnInputTuple_{batch}$ are therefore normalized between 0 and 1.
% The complete input for the policy NN of the batch variant becomes a set of $\nnSampleSize$ batches $\nnInputBatchBatch$: $\nnInputBatch = \{\nnInputBatchBatch_1, …, \nnInputBatchBatch_\nnSampleSize\}$.

As for the single variant, we select a good initial guess of $\nnSampleSize$ batches during the pre-sampling phase too, as we can not provide all batches at once for the policy NN.
The heuristic selects potentially good batches based on three objectives: first, we select batches, whose samples are the furthest apart from each other:
\begin{align}
	\nnInputBatchBatch_{dist} = \argmax_{\randomActionSet} \left(\sum_{x_i \in \randomActionSet} \sum_{x_j \in \randomActionSet} d(x_i, x_j)\right)
	\label{eq:b_dist}
\end{align}
Then we select batches, whose samples have the highest average uncertainty $\uncFunc_1$:
\begin{align}
	\nnInputBatchBatch_{\uncFunc} = \argmax_{\randomActionSet} \left(\sum_{x_i \in \randomActionSet}\uncFunc_1(x_i)\right)
	\label{eq:b_unc}
\end{align}
% Both objectives are computed for $\preSamplingIterations$ random possible batches, instead of all possible subsets of $\unlabeledSet$.
Both $\argmax$ can return multiple batches.
The final set of potentially good batches consists of $\frac{2}{5}\nnSampleSize$ batches from Equation~\ref{eq:b_dist}, $\frac{2}{5}\nnSampleSize$ batches from Equation~\ref{eq:b_unc}, and $\frac{1}{5}\nnSampleSize$ random batches in case the two previous objectives did not select good batches.
%and is the return value of $\Call{PreSampling}{}$ for the batch variant.
%
%
%
\subsection{Policy NN Output Encoding}
\label{sec:action_space}
The output encoding of the final layer of the policy NN indicates which action to take.
For both the single and the batch variant of \ImitAL{} we have a final softmax layer of $\nnSampleSize$ output neurons, each per possible action.
For the single variant, the $\batchSize$ highest output neurons indicate the samples for the unlabeled query $\unlabeledQuery$.
The batch variant indicates with the highest output neuron directly the index of the pre-sampled batch that represents the unlabeled query $\unlabeledQuery$.
\section{Training Implementation Details}
\label{sec:train_imital}
Most learned AL methods~\cite{PAL,WoodwardandFinn,ALIL,LAL-RL,Bachman,Pang_single} are frameworks on how to learn an AL strategy.
Given domain-related pre-labeled datasets, one can train the AL strategy on them, and apply it afterward on an unlabeled dataset of a similar domain.
In contrast, we provide an already trained AL strategy, which works directly without the aforementioned required transfer learning step.
Under the assumption that a large set of diverse synthetic datasets can be generated, our trained AL strategy seems to be universally applicable.
We present in Section~\ref{sec:train_impl_details} the implementation details used for training \ImitAL{}, and in Section~\ref{sec:synthetic_datasets} the details of the used generated synthetic datasets.
%
%,
%
\subsection{Implementation Details}
\label{sec:train_impl_details}
\begin{table}[t]
    \small
	\caption{Used Hyperparameters for the policy NN of \ImitAL{}}
	\centering
    \addtolength{\tabcolsep}{-6pt}
	\begin{tabular}{lrr}
		\toprule
		Hyperparameter             & \ImitAL{} Single   & \ImitAL Batch      \\
		\midrule
		\#input neurons            & $5*\nnSampleSize$  & $3* \nnSampleSize$ \\
		\#output neurons           & $\nnSampleSize$    & $\nnSampleSize$    \\
		\#hidden layers            & 2                  & 3                  \\
		\#hidden neuron layer size & 1,100              & 900                \\
		\#NN batch size            & 128                & 128                \\
        loss function              & mean squared error (MSE) & MSE \\
		activation function        & elu                & elu                \\
		dropout rate               & 0.2                & 0.2                \\
		initial learning rate      & 0.001              & 0.001              \\
		max \#epochs               & 1,000              & 1,000              \\
		early stopping             & True               & True               \\
		% Target Output         & Regression         & Regression         \\
		\bottomrule
	\end{tabular}
	\label{tab:hyperparams}
\end{table}
We generated $\AmountOfSDs = 30,000$ synthetic datasets to train \ImitAL{}.
More training data did not seem to improve the quality of the learned policy NN.
As starting point one random sample per class was labeled to initially train the AL learner model $\learner$.
Each synthetic dataset was used for $\AmountOfTrainingALCycles = 10$ AL cycles. %We set the amount of AL cycles to use each synthetic dataset $\AmountOfTrainingALCycles$ to 10.
Given a fixed computational budget one has to choose between more synthetic datasets or more AL cycles.
As the influence of good AL query selection, and therefore a good imitation learning expert demonstration policy is more prominent for early AL cycles, we set $\AmountOfTrainingALCycles$ to a seemingly rather small number of 10.
But according to our experience, it is better to generate more synthetic datasets and observe more early AL cycles than to concentrate on fewer datasets with more cycles.
Therefore, in total, we generated 300,000 pairs of synthetic-dataset-expert actions or training samples for the NN.

The batch size $\batchSize$ was set to 5.
We set the number of input samples for the NN $\nnSampleSize = 20$.
Larger values mean of course better evaluation performance but at the cost of computation performance.
During the pre-sampling phase randomly $\preSamplingIterations$-times possible actions are drawn, and then out of those, according to the heuristic, the best actions are taken.
The parameter $\preSamplingIterations$ was set to 10 for the single variant, and 750 for the batch variant.
Both values worked best in our experiments.
Additionally, our experiment showed that the heuristic of the pre-sampling phase is not needed anymore after training for the single variant, but still for the batch variant.
For the single variant, we, therefore, recommend simply to select $\nnSampleSize$ random samples as input.
With the values for $\nnSampleSize$ and $\batchSize$ the policy NN selects the 5 best samples out of possible 20.
Thus, the single variant of \ImitAL{} works in a batch-mode aware setting.

The synthetic datasets were normalized using Min-Max scaling to the range between 0 and 1 as pre-processing.
For the normalization of the distance part of the input encoding of the batch-variant, the maximum distance $\maxLength$ in the vector space is needed.
Due to the min-max scaling it becomes $\sqrt{(\#features)}$.
As distance metric, we used for training \ImitAL{} the Euclidean distance, as it proved best for the synthetic datasets.
While applying the trained strategy later to large real-world datasets we noticed that the cosine distance works better for large datasets when using the single variant, so we recommend adjusting the distance metric based on the size of the vector space.

Since AL is applied in an early stage of ML projects, we chose random forest classifier as a simple AL model as learner $\learner$.
They are robust and work well for many dataset domains, without a lot of parameters to fine-tune, which one often does not know at that early project stage.
Additionally, they are fast to train and offer a sound calculation of the probability $P_\learner$ using the mean predicted class probabilities of the trees in the forest, whereas the class probability of the trees is defined by the fraction of samples of the same class in a leaf.
We used the random forest classifier implementation in the widely popular Python library scikit-learn~\cite{scikit-learn}, which is a slightly adapted version of~\cite{Breiman_RF}.

% Two versions of the training target output were tested - one where the final output layer directly predicted the expected resulting accuracy for each unlabeled sample, if one would label this sample, resulting in a regression problem, and one, where for the best $\batchSize$ samples the target output was $1$, and for the other $\nnSampleSize - \batchSize$ neurons the output target was $0$, a classification problem.
The gathered state-action pairs were split into 70\% training and 30\% test data.
We performed a non-exhaustive random search for selecting the best hyperparameters for the policy NN, which are displayed in Table~\ref{tab:hyperparams}.
The final NN of the single variant consists of an input layer with $5*\nnSampleSize$-input neurons, two hidden layers with 1,100 neurons each, \emph{elu} as activation function, and a final layer with $\nnSampleSize$-output neurons and a $sigmoid$ activation function.
The batch variant uses an input layer with $3*\nnSampleSize$-input neurons and three hidden layers with 900 neurons each.
To prevent overfitting while training a dropout rate of 0.2 was used.
The optimizer during training is \emph{RMSprop}, the loss function \emph{MeanSquaredError}, the batch size is 128, and the learning rate initially 0.001, but got reduced with factor 10 on plateaus.
% The number of training epochs was capped at 10,000 but stopped early if the validation loss was plateauing.
If the validation loss was plateauing, first the learning rate got reduced by factor 10, and after that, the training stopped early.
To prevent the policy NN from learning the positional ordering of the inputs we also tested out generating new training samples by permutating the state-action pairs of the NN.
But as this had no real influence on the training loss except for much longer training times, we can not recommend this approach.
Apparently the huge amount of used synthetic datasets provides already enough training data to generalize from positional information.
The training was conducted using four CPU cores and 20 GB of RAM on an HPC cluster running Linux Kernel 3.10 and Python 3.6.8.
In total {\raise.17ex\hbox{$\scriptstyle\sim$}}200,000 CPU-hours were needed for all experiments conducted for this paper, including testing out different input encodings and training the final versions of \ImitAL{}.
\subsection{Synthetic Datasets}
\label{sec:synthetic_datasets}
The synthetic datasets used for generating training data are an important part of the training setup of \ImitAL{}.
The better they resemble possible real-world datasets, the more universally applicable our trained policy NN is.
Table~\ref{tab:synth_datasets} lists the range of parameters and random distributions used while generating the synthetic datasets.
\begin{table}[t]
    \small
	\caption{Generation parameters for synthetic datasets}
	\centering
	\begin{tabular}{lr}
		\toprule
		Parameter                 & Distribution and range \\
		\midrule
		\#samples                 & uniform(100, 5,000)    \\
		\#features                & uniform(2, 100)        \\
		\#classes                 & uniform(2, 10)         \\
		\#clusters per class      & uniform(1, 10)         \\
		\%class weights per class & dirichlet(0, 100)      \\
		\% of random label noise  & pareto(0, 100)         \\
		class separability        & uniform(0,10)          \\
		\bottomrule
	\end{tabular}
	\label{tab:synth_datasets}
\end{table}
We use the implementation of the algorithm by~\cite{guyon2003design} in \emph{scikit-learn}~\cite{scikit-learn} for generating the synthetic datasets.
The number of generated features is split up randomly between redundant, repeated, and informative features.
Essentially, with the number of informative features as the dimension size, random hypercubes are generated.
Then, clusters of normally distributed points are scattered in the hypercubes.
The length of the hypercube sides is $2^{class\ separability}$, meaning larger values of $class\ separability$ result in more spread out samples and therefore harder classification tasks.
The class weight parameter controls the ratio between the number of samples between the classes.
We explicitly decided to not only focus on binary classification problems but to set the number of classes with a maximum of 10, as the more classes exist, the harder is the labeling process, and the more useful is AL in general.
We also did not decide to create very large datasets with many samples as this would firstly, drastically extend the training duration, and, secondly, AL strategies should be able to easily scale up from small to large datasets.
%
%
%
% \subsection{old content not yet moved somewhere meaningful}
% In contrast to the single variant, the pre-sampling phase for the batch variant is also needed during the application of the trained network.
% Since we can calculate a very large amount of different synthetic datasets and thereby a lot of demonstrative expert actions, and no real "erroneous" actions can happen by labeling a non-optimal sample, we used the generated pairs of synthetic dataset and expert actions directly as standard supervised training data.
% The batch variant calculates the accuracy after adding the whole batch.

% Inspired by the approach by~\cite{MAIKPAPIER} and ~\cite{LAL} we used a large amount of synthetic datasets (SD) with various parameter ranges.
%
%
%
\section{Evaluation}
\label{sec:evaluation}
\begin{table}[t]
    \small
	\caption{Overview of evaluation datasets}
	\centering
	\begin{tabular}{lrrrr}
		\toprule
		dataset    & \#samples & \#features & \#classes & \#AL cycles \\
		\midrule
		australian & 690       & 14         & 2         & 50          \\
		CIFAR-10   & 10,000    & 3,072      & 10        & 1,000       \\
		diabetes   & 768       & 8          & 2         & 50          \\
		DWTC       & 5,777     & 227        & 4         & 50          \\
		EMNIST     & 116,323   & 784        & 62        & 1,000       \\
		fertility  & 100       & 20         & 2         & 50          \\
		flags      & 194       & 48         & 8         & 50          \\
		german     & 1,000     & 24         & 2         & 50          \\
		glass      & 214       & 9          & 6         & 50          \\
		haberman   & 306       & 3          & 2         & 50          \\
		heart      & 303       & 13         & 2         & 50          \\
		ionosphere & 351       & 34         & 2         & 50          \\
		olivetti   & 400       & 4,096      & 40        & 50          \\
		planning   & 182       & 12         & 2         & 50          \\
		zoo        & 101       & 16         & 7         & 50          \\
		\bottomrule
	\end{tabular}
	\label{tab:datasets}
\end{table}
\begin{table*}[t]
    \small
	\caption{AUC-F1-scores (\%) for different AL query strategies, mean for 100 repeated experiments each, including the ranks and the ranked mean. Empty cells indicate no calculable results within the maximum runtime window of seven days.}
	\addtolength{\tabcolsep}{-5.30pt}
	\centering
	\begin{tabularx}{\linewidth}{L{1.1cm}rrrrrrrrrrrrrr}
\toprule
    & NN Single   & NN Batch   & MM        & QBC        & LC         & GD         & BMDR       & Ent        & Rand       & LAL        & SPAL       & EER        & QUIRE     \\
\midrule
 australian &\fontseries{b}\selectfont{85.4 (0)  }& 85.3 (4)  & 85.4 (2) & 85.3 (3)  & 85.4 (1)  & 85.2 (5)  & 85.1 (7)  & 85.4 (2)  & 84.9 (8)  & 85.2 (6)  & 84.3 (10) & 84.6 (9)  & 76.2 (11)\\
 \mbox{CIFAR-10}   &\fontseries{b}\selectfont{37.0 (0)  }& 36.9 (1)  & 36.9 (2) & 36.0 (5)  & 35.9 (6)  & 36.3 (4)  &   & 35.7 (7)  & 36.8 (3)  & 34.9 (8)  &   &   &  \\
 diabetes   & 74.3 (4)   & 74.5 (2)  & 74.3 (5) & 74.4 (3)  & 74.3 (6)  &\fontseries{b}\selectfont{75.0 (0) }& 74.7 (1)  & 74.3 (5)  & 74.2 (8)  & 74.3 (7)  & 73.9 (10) & 74.2 (9)  & 66.5 (11)\\
 DWTC       &\fontseries{b}\selectfont{74.9 (0)  }& 74.1 (2)  & 74.7 (1) & 71.7 (6)  & 71.0 (7)  & 65.1 (9)  &  & 68.5 (8)  & 72.9 (3)  & 72.2 (5)  &  & 72.8 (4)  & 53.4 (10)\\
 EMNIST     & 69.1 (1)   & 68.6 (2)  &\fontseries{b}\selectfont{69.5 (0)}& 63.4 (4)  & 61.6 (5)  & 43.8 (7)  &   & 59.2 (6)  & 66.1 (3)  &   &   &   &  \\
 fertility  & 88.5 (3)   & 88.3 (4)  & 88.0 (8) & 88.2 (6)  & 88.0 (8)  & 88.3 (5)  & 88.6 (2)  & 88.0 (9)  & 87.4 (10) & 88.0 (7)  & 88.6 (1)  & 86.6 (11) &\fontseries{b}\selectfont{89.1 (0)} \\
 flags     &\fontseries{b}\selectfont{57.7 (0)  }& 56.4 (5)  & 56.6 (4) & 55.9 (9)  & 55.4 (10) & 57.0 (2)  & 57.4 (1)  & 54.7 (11) & 56.2 (6)  & 55.9 (8)  & 56.9 (3)  & 56.0 (7)  & 50.4 (12)\\
 german     & 75.7 (2)   & 75.6 (3)  & 75.4 (9) & 75.5 (6)  & 75.4 (10) & 76.2 (1)  & 75.6 (5)  & 75.4 (11) & 75.6 (4)  & 75.5 (7)  &\fontseries{b}\selectfont{76.5 (0) }& 75.4 (8)  & 70.9 (12)\\
 glass      &\fontseries{b}\selectfont{69.0 (0)  }& 68.1 (2)  & 68.6 (1) & 67.5 (6)  & 68.0 (3)  & 66.4 (8)  & 67.6 (5)  & 66.3 (9)  & 67.6 (4)  & 66.0 (10) & 61.7 (11) & 67.2 (7)  & 48.2 (12)\\
 haberman   & 72.9 (4)   & 72.3 (5)  & 72.9 (3) &\fontseries{b}\selectfont{73.3 (0) }& 72.9 (1)  & 71.6 (8)  & 71.6 (9)  & 72.9 (2)  & 71.5 (10) & 72.3 (6)  &  & 72.0 (7)  & 63.2 (11)\\
 heart      &\fontseries{b}\selectfont{79.4 (0)  }& 79.4 (1)  & 79.1 (7) & 79.1 (9)  & 79.1 (6)  & 79.2 (4)  & 79.3 (2)  & 79.1 (7)  & 79.1 (8)  & 79.3 (3)  & 78.6 (10) & 79.1 (5)  & 74.7 (11)\\
 ionosphere &\fontseries{b}\selectfont{90.3 (0)  }& 89.3 (9)  & 90.0 (2) & 89.9 (4)  & 90.0 (1)  & 89.6 (6)  & 89.8 (5)  & 90.0 (1)  & 89.6 (7)  & 89.9 (3)  & 83.2 (10) & 89.4 (8)  & 53.7 (11)\\
 olivetti   & 72.2 (1)   & 70.4 (5)  &\fontseries{b}\selectfont{72.3 (0)}& 72.1 (3)  & 72.1 (2)  & 65.5 (10) & 64.3 (11) & 70.7 (4)  & 66.1 (7)  & 65.6 (9)  & 66.5 (6)  & 65.7 (8)  & 51.2 (12)\\
 planning   & 72.3 (3)   &\fontseries{b}\selectfont{74.0 (0) }& 72.1 (4) & 72.7 (2)  & 72.1 (4)  & 69.6 (8)  & 68.9 (9)  & 72.1 (5)  & 68.4 (10) & 71.5 (7)  & 73.6 (1)  & 68.2 (11) & 71.8 (6) \\
 zoo        & 93.4 (1)   &\fontseries{b}\selectfont{93.5 (0) }& 93.0 (4) & 92.5 (10) & 92.9 (7)  & 92.7 (9)  & 93.3 (2)  & 92.8 (8)  & 93.0 (5)  & 92.3 (11) & 92.2 (12) & 92.9 (6)  & 93.2 (3) \\
 \midrule 
 mean \%    &\fontseries{b}\selectfont{74.2 (0)  }& 73.8 (2)  & 73.9 (1) & 73.2 (3)  & 72.9 (4)  & 70.8 (7)  & 61.1 (10) & 72.3 (6)  & 72.6 (5)  & 68.2 (8)  & 55.7 (12) & 65.6 (9)  & 57.5 (11)\\
 mean (r)   & \fontseries{b}\selectfont{1.27}        & 3.00       & 3.47      & 5.07       & 5.13       & 5.73       & 5.80       & 6.33       & 6.40       & 7.00       & 7.60       & 7.80       & 9.27      \\
\bottomrule
\end{tabularx}

	\label{tab:alipy_f1}
\end{table*}
The evaluation consists of an extensive comparison between \ImitAL{} and 10 other state-of-the-art AL query strategies on a total of 15 diverse datasets.
We also take a look at the runtime performance in addition to a purely qualitative analysis. %, and try to understand what type of heuristic was being learned by the NN.
\subsection{Experiment Setup Details}
\label{sec:exp_setup_details}
For evaluation, we selected 15 common publicly available datasets representing a diverse field of domains to test our trained strategy on.
The datasets are mostly small and medium-sized ones from the UCI Machine Learning Repository~\cite{uci}, similar to the evaluations of~\cite{Pang_single,LAL-RL,QUIRE,ALBL,BMDR}.
In our experience, AL strategies that perform well on small datasets also perform well on large datasets.
Due to much longer experimentation runtimes, the majority of our selected datasets are smaller ones.
Additionally, DWTC~\cite{DWTC} is used as an example for a medium-sized dataset with more than two classes from the domain of table classification.
Olivetti~\cite{olivetti} has a large vector space, CIFAR-10~\cite{cifar10}, and the test set of EMNIST~\cite{emnist}, are added as representatives of large datasets with many samples.
Table~\ref{tab:datasets} contains an overview of the used datasets.
The experiments were repeated 100 times with different random starting points.
Each AL cycle started with the least possible number of needed labels, which is one labeled sample per class.
The batch size $\batchSize$ was set to 5 samples.
After 50 AL cycles, or when all data was labeled, the AL process was stopped for most of the datasets, as at this point already a clear distinction could be made between the competing AL query strategies.
For the two large datasets, EMNIST and CIFAR-10, the number of AL cycles was extended to 1,000.
All datasets were normalized using Min-Max scaling to the range of 0 and 1.
The train-test split was always 50\% - 50\% and next to the labeled start set also randomly chosen per experiment.
We used again random forest classifier as learner model $\learner$.
On often has not enough knowledge about elaborate and specialized classification models at that early stage of machine learning projects when applying AL.
Other learners such as neural networks or support vector machines would have been possible as well, since \ImitAL{} does not depend on any specific properties of the AL learner model.
Experiments using neural networks as learner provided generally similar results to the ones presented using random forest classifier, but with larger runtimes.
The distance metric was changed to cosine for the single variant for DWTC, CIFAR-10, EMNIST, and olivetti due to the large vector spaces.
The experiments were conducted on the same HPC as used for training the policy NN: four CPU cores, 20 GB of RAM, and a maximum runtime window of seven days.

Many other AL papers base their evaluation primarily on the interpretation of comparing individual AL learning curves with each other.
However, since we repeated our experiments per dataset 100 times with different starting points, we could not use the same procedure as we ended up with 100 learning curves per dataset and AL query strategy.
As the evaluation metric, we used, therefore, the \emph{area under the F1-learning curve} (AUC-F1), similar to the evaluation metric of the Active Learning challenge from Guyon et al.~\cite{ALChallenge} .
This metric condenses the complete learning curve to a single number.
Given a standard AL curve, where after each AL cycle $t$ the F1-score $F_t$ was measured, the AUC-F1 is the area under the learning curve using the trapezoidal rule, normalized to the range between 0 and 1:
\begin{align}
	\Call{AUC-F1}{F_1, \dots, F_n} = \frac{1}{2n-2} \left(F_1 + \sum_{t=2}^{n-1} 2*F_t + F_n \right)
	\label{eq:aucf1}
\end{align}
An optimal AL query strategy would go straight to an F1-Score of 1.0, and continuing on the top, resulting in a rectangle with an area of 1.
The worst AL strategy has a complete flat AL curve at a constant F1-Score of 0.0 and an area of 0.
The AUC-F1 score rewards therefore early good decisions more than later ones.
% The learning curve is the F1-score measured after each AL cycle.
We chose the F1-score as it works well for imbalanced datasets with more than two classes, and the area under the curve as it punishes bad samples in early AL cycles.
\subsection{Competitors}
We compare \ImitAL{} with 10 different state-of-the-art AL query strategies using the implementations in ALiPy~\cite{ALiPy}.
The first three query strategies are the uncertainty variants Max-Margin (MM)~\cite{mm_sampling}, Least Confidence (LC)~\cite{lc_sampling}, and Entropy (Ent)~\cite{ent_sampling}, which all rely purely on the probabilistic output of the learner to select the most uncertain samples.
Another standard AL query strategy is the bagging variant of Query-by-Committee (QBC)~\cite{qbc_sampling_bagging}, where the goal is to minimize the set of hypotheses that are consistent with the current labeled set.
Expected Error Reduction (EER)~\cite{EER} maximizes the expected \emph{mutual information} of the query samples and the unlabeled set $\unlabeledSet$.
Querying Informative and Representative Examples (QUIRE)~\cite{QUIRE} solves AL using a min-max view of AL.
Graph Density (GD)~\cite{GD} uses a connectivity graph to sample from all dense regions evenly distributed.
Query Discriminative and Representative Samples for Batch Mode Active Learning (BMDR)~\cite{BMDR} formulate AL as an ERM risk-bound problem.
Learning Active Learning (LAL)~\cite{LAL} trains a Random Forest Regressor from simple two-dimensional synthetic datasets, and predicts the expected error reduction.
The most recent method is Self-Paced Active Learning (SPAL) ~\cite{SPAL}, where AL is formulated as an optimization problem with the goal to query easy samples first, and hard ones later when the learned model is more robust.
And finally, we include the random selection (Rand) of samples as another baseline strategy.
\subsection{Results}
% \begin{figure}[t]
%     \begin{center}
%         \includegraphics{figures/iso.pdf}
%         \caption{Isomap of column-vectors for Table~\ref{tab:alipy_f1} to understand the learned strategy by \ImitAL{}}
%         \label{fig:iso}
%     \end{center}
% \end{figure}%
Table~\ref{tab:alipy_f1} shows the results for all datasets and AL query strategies.
Empty cells occur because the query strategies presented could not finish the experiment within the maximum runtime of seven days.
The first two columns show the result of \ImitAL{} in the single and batch variant.
As the experiments were repeated 100 times per dataset with different starting points, the displayed value per dataset is the arithmetic mean.
Due to the used AUC-F1 metric each displayed number represents 100 complete learning curves and the whole table 18,300 learning curves without the empty cells. %100*15*13 - 12*100
The second to last row shows the mean across all datasets.
Additionally, the last the last row shows the average of the ranks, which are shown in parentheses after the percentages.
As the main goal for \ImitAL{} is to learn an universally applicable query strategy it needs to perform well across all datasets, and not just stand out drastically for specific ones.
Note that the displayed percentages are rounded, but the ranks are computed on the complete numbers, which can lead to different ranks for otherwise equally rounded displayed percentages.
It can be seen that some strategies like SPAL, GD, QBC, or QUIRE perform very well for specific datasets, but at the same time very poorly for almost all other datasets.
Most of the advanced AL query strategies perform often close to or worse than Rand for most datasets.
To be fair, several of those are designed primarily for binary classification problems.
Quite good results come still from the simple and early proposed AL query strategies MM, QBC, and LC.
The second-best strategy is MM.
First, it is interesting that MM is far better than the close variant LC, and again better than Ent.
But still, even though MM performs best for two datasets, it also performs very poorly on other datasets such as \emph{fertility}, \emph{german}, or \emph{heart}.
As shown before~\cite{QUIRE,GD,LAL}, purely uncertainty-based strategies are prone to sampling bias and therefore unable to deal with XOR-like datasets, which is probably the cause for the shortcoming on these datasets.
The single variant of our approach \ImitAL{} has the lead by a small margin on the percentage mean, and by a large margin on the ranked mean.
In addition to that, it is the best strategy for 7 datasets, whereas the batch variant or MM excel on two.
From the point of view of a universally well-performing strategy, the rank is the more interesting metric, as it does not overvalue single spikes in performance for specific datasets.
Even though the single variant of \ImitAL{} is not always the best strategy, it still receives always high ranks even on its worse-performing datasets, lacking a dataset where it performs poorly.
It is also the only AL query strategy, which never performs worse than random.
It may appear so that the single variant of \ImitAL{} is most challenged on datasets with a large number of features in comparison to the number of samples.
But if one would limit Table~\ref{tab:alipy_f1} to only those datasets that have the highest ratio between the number of features and the number of samples (\emph{olivetti}, \emph{CIFAR-10}, \emph{flags}, \emph{fertility}, and \emph{zoo}) the mean rank still shows that the single variant of \ImitAL{} is better than all other compared approaches.

The batch variant is rank-wise still better than MM, but worse on the percentage mean.
Still, it performs even better than the single variant on two datasets, \emph{planning} and \emph{zoo}, which indicates that the input encoding of the single variant is not optimal yet.
Interestingly, the restriction due to the fixed size input of the policy NN from \ImitAL{} has therefore for both variants no significant harming influence on the performance.
In contrast, all other AL query strategies consider always all unlabeled samples for a decision.

\subsection{Performance}
\begin{figure}[t]
	\begin{center}
		\includegraphics{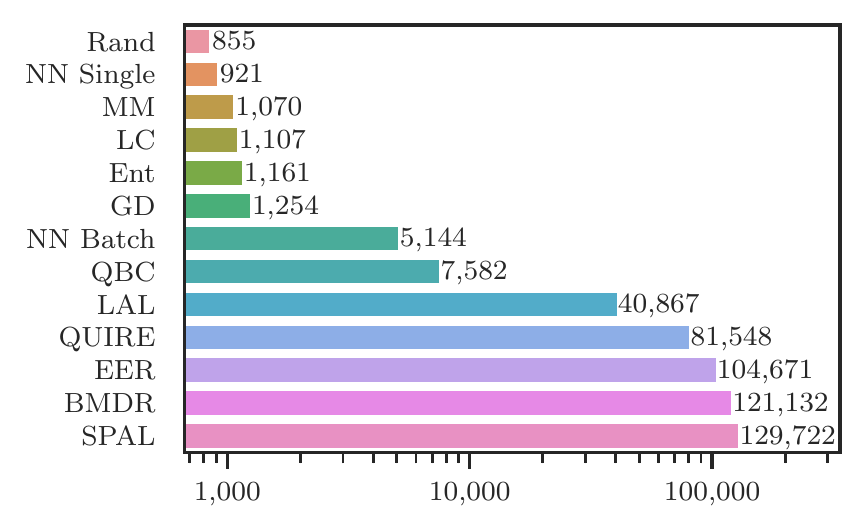}
		\caption{Average runtime duration in seconds per complete AL experiment, with timeout duration for experiments lasting longer than 7 days}%those who did not finish within the timeout are included with the timeout}
		\label{fig:performance}
	\end{center}
\end{figure}
The second goal for \ImitAL{} is to provide a superior small runtime, which should make AL more easily applicable to large-scale datasets.
For that, we calculated the mean of the runtime of the complete AL experiments the same way as the AUC-F1-scores in Table~\ref{tab:alipy_f1}, which is shown in Figure~\ref{fig:performance}.
It is not surprising that the random query strategy is the fastest of all query strategies, as this strategy does not depend on the number of unlabeled data.
Almost as fast is single \ImitAL{}, which always considers a fixed number of samples $\nnSampleSize$ as input encoding for the policy NN.
For the batch variant, $\preSamplingIterations$-times $\nnSampleSize$ samples are considered during the pre-sampling phase.
As the heuristic for the pre-sampling phase of the batch variant is computationally expensive and needs a lot of iterations before finding a good set of potential batch candidates, it performs worse than the single variant, which can work well on arbitrary random input.
All other query strategies need at least one full pass over all unlabeled samples before a decision can be made.
Techniques like down-sampling can of course circumvent a full pass, but come at the cost of a performance loss.
The difference is reduced if one would make the performance analysis only on small datasets.
% This difference in the full-pass becomes of course less if one would make the performance analysis only on small datasets.
For these, the uncertainty strategies are slightly faster than the single variant of \ImitAL{}.
But especially on the very large dataset EMNIST it becomes clear, what difference a fixed size input can make as opposed to a full-pass.
As a result, the single variant of \ImitAL{} is on average the fastest real AL query strategy.
% The reason why the single variant of \ImitAL{} performs even better than the simple uncertainty
Other techniques like BMDR, SPAL, EER, or QUIRE need significantly more time than most other strategies, which hinders their application in real-world AL scenarios.
BMDR and SPAL encode the AL problem as a quadratic optimization problem, which explains the exploding runtime performance for large datasets.
It appears that the provided implementation of LAL trains the query strategy model at the start of each AL experiment from scratch.
This could potentially be further optimized by doing the training separately once for all experiments, as it is being done with \ImitAL{}.
% EER is known to 
% EER
% The batch variant of \ImitAL{} needs a lot more compute-intensive pre-sampling iterations to work well than the single variant, but still performs better than a standard QBC approach.
% Nevertheless, if performance is important while labeling we would highly recommend to use the single variant instead.

In conclusion, we would recommend using the single variant of \ImitAL{}, firstly with regard to the best results for the AUC-F1-scores, percentage, and rank-wise, and secondly because of the general fastest runtime of all query strategies.
\section{Related Work}
\label{sec:related_work}
Besides traditional AL strategies like~\cite{lc_sampling,mm_sampling,EER,GD,qbc_sampling,BMDR,SPAL} the field of learning AL query strategies has emerged in the past few years.
Instead of relying on heuristics or solving optimization problems, they can learn new strategies based on training on already labeled datasets.

ALBL~\cite{ALBL} encodes in a QBC-like approach the AL query strategy as a multi-armed bandit problem, where they select between the four strategies Rand, LC, QUIRE, and a distance-based one.
To function properly a reward for the given dataset is needed during the application of their approach as feedback.
They propose to use a dedicated test set for feedback, which can in practice rarely be taken for granted.
Their tests on computing the reward on during AL acquired labels showed that a training bias can occur resulting in poor performance.
% This only works well, when a separate test set is available, as the reward computed purely on the given labels during AL can perform poorly because of training bias.
LAL~\cite{LAL} uses a random forest classifier as the learner, and the input for their learned model are statistics of the state of the inner decision trees from the learner for the unlabeled samples.
% Based on this they predict the expected error reduction.

Most of the other methods rely on Reinforcement Learning (RL) to train the AL query strategy.
A general property that distinguishes the RL and imitation learning-based methods is the type of their learning-to-rank approach.
\cite{WoodwardandFinn,PAL,LAL-RL,ALIL} all use a pointwise approach, where their strategy gets as input one sample at a time. %decides per sample individually if it should be labeled or not.
They all need to incorporate the current overall state of $\labeledSet$ and $\unlabeledSet$ into their input encodings to make the decision work on a per-sample basis.
\cite{LAL,Bachman,Pang_single}, as well as \ImitAL{}, use the listwise approach instead, where the strategy gets a list of unlabeled samples at once as input.
This also has the benefit of a batch-aware AL setting.

The most distinctive characteristic of the RL and imitation learning-based approaches is the input encoding.
\cite{Bachman} uses the cosine similarities to $\labeledSet$ and $\unlabeledSet$ as well as the uncertainty of the learner.
\cite{ALIL} incorporates directly the feature vectors, the true labels for $\labeledSet$, and the predicted label of the current pointwise sample.
They also use imitation learning instead of RL with the same future roll-out of the AL cycle as the expert as we propose for \ImitAL{}.
\cite{PAL} adds to their state additionally the uncertainty values for $\unlabeledSet$ and the feature vectors.
This has the limitation that the trained strategies only work on datasets having the same feature space as the datasets used in training.
\cite{Pang_single} bypasses this restriction by using an extra NN which maps the feature space to a fixed size embedding.
Therefore they are, at the cost of complexity of an additional layer, independent of the feature space, but can still use the feature vectors in their vector space.
A big limitation is that their embedding currently only works for binary classification.
Additionally, they add distance and uncertainty information to their state.
\cite{LAL-RL} does not add the vector space into their input encoding.
To incorporate the current state for their pointwise approach, they use the learners' confidence on a fixed size set of unlabeled samples.
Further, they use the average distances to all samples from $\labeledSet$ and $\unlabeledSet$ each as well as the predicted class to encode the pointwise action.

Our input encoding uses uncertainty, distance, and for the batch variant a disagreement score among the predicted batch labels.
Due to our listwise approach, we only need to add this information for the to-rank samples to our state, and not $\labeledSet$ or $\unlabeledSet$.
We do explicitly not add the feature vectors, as this limits the transferability of the trained strategy to new datasets, which contradicts our goal of a universal AL strategy.

Most of the works rely on training on domain-related datasets before using their strategy.
This prevents an already trained AL query strategy from being easily reusable on new domains.
\cite{LAL} bypasses this by training on simple synthetic datasets, but due to their simple nature, they still recommend training on real-world datasets too.
Our approach of using a large number of purely random and diverse synthetic datasets during training gives \ImitAL{} the benefit of not needing an explicit prior training phase on domain-related datasets.

The necessary pre-training of many related works as well as the often not publicly available code prevented them from being included in our evaluation.
% \todo{in text einarbeiten?}
% On a final note an appeal to fellow AL researchers: Learned AL query strategies have reached a point of complexity that makes it nearly impossible to reimplement all of them for an evaluation against newly developed methods.
% On top, a large number of parameters of AL experiments like batch size, stopping criteria, evaluation metric, train/test split, learner classification model, or starting samples, make it hard to simply compare results even if they were conducted on the same datasets.
% We urge therefore everyone to make their code not only publicly available but also to provide enough information about how to run the code.
% 
% \cite{WoodwardandFinn,Bachman} also suffer from the problem that they are not agnostic of the learner model, which prevents them to be adapted to the given dataset domain.
% \cite{Zhang_weak_strong}
% es gibt 2 oracles: strong but expensive + weak but cheap labeller, wie kann man mit beiden arbeiten?
%
%
%
\section{Conclusion}
\label{sec:conclusion}
We presented a novel approach of training a near universally applicable AL query strategy on purely synthetic datasets, by encoding AL as a listwise learning-to-rank problem.
For training, we chose imitation learning, as it is cheap to generate a huge amount of training data when relying on synthetic datasets.
Our evaluation showed two properties of \ImitAL{}: first, it works consistently well across various numbers of datasets and is not limited to a specific domain or learner model, outperforming other state-of-the-art AL query strategies, and, second, it is reasonably faster than its competitors even on large-scale datasets.
In the future, we want to include more requirements of large machine learning projects into the state-encoding of \ImitAL{} to make it more applicable.
Large multi-label classification target hierarchies are often very hard to label but propose new challenges for AL.
Additionally, different label often have varying costs and should be treated accordingly by the AL query strategy~\cite{Donmez}.
On a similar note, given a lot labels by multiple noisy sources like crowdsourcing relabeling using AL becomes important~\cite{Zhao}.
% Often, the labeling cost varies among the unlabeled samples: some samples can only be labeled by expensive experts, while others can be outsourced to cheap labor~\cite{Donmez}.
% On a similar note, given a lot labels by multiple sources like from crowdsourcing~\cite{Zhao}, what are the labels that are potentially wrong and should be relabeled using AL?
% Besides complete human labeling, weak supervision~\cite{Snorkel} should be includedcan be considered when deciding what to label by the domain experts, and when to take the automatic weakly generated label.
% Classification problems with thousands of different potentially multi-label targets are often especially hard to label and should be considered when developing an AL strategy as well.

We only trained \ImitAL{} on purely synthetic datasets.
It is also possible to retrain it first using domain-specific datasets in a transfer learning setting before applying.
The effectiveness of this approach remains open to study.
Other possible concrete improvements of \ImitAL{} could be a long short-term memory NN which remembers the state of the previous AL cycles or more elaborate imitation learning strategies like DAGGER~\cite{DAGGER}.
% like variable labeling costs, multiple noisy oracles, or changing classification target classes due to transfer learning.

Looking back to Figure~\ref{fig:optimal} there is a lot of room for improvements from current state-of-the-art AL query strategies compared to an optimal strategy.
Although \ImitAL{} performed better than the best baseline uncertainty max-margin, the improvement is not of the possible order of magnitude shown in Figure~\ref{fig:optimal}, which further indicates that AL query strategy research is still an interesting open research questions.
% This further indicates that research on AL query strategies

\subsection*{Acknowledgment}
	This research and development project is funded by the German Federal Ministry of Education and Research (BMBF) and the European Social Funds (ESF) within the "Innovations for Tomorrow's Production, Services, and Work" Program (funding number 02L18B561) and implemented by the Project Management Agency Karlsruhe (PTKA). The author is responsible for the content of this publication.

\bibliographystyle{plain}
\bibliography{paper}

\end{document}